\documentclass[runningheads]{llncs}
\usepackage{graphicx}
\usepackage{amsmath,amssymb} % define this before the line numbering.
\usepackage{booktabs}
\usepackage{makecell}
\usepackage{color}
\usepackage{multirow}
\usepackage{multicol}

\usepackage[width=122mm,left=12mm,paperwidth=146mm,height=193mm,top=12mm,paperheight=217mm]{geometry}

\begin{document}
% \renewcommand\thelinenumber{\color[rgb]{0.2,0.5,0.8}\normalfont\sffamily\scriptsize\arabic{linenumber}\color[rgb]{0,0,0}}
% \renewcommand\makeLineNumber {\hss\thelinenumber\ \hspace{6mm} \rlap{\hskip\textwidth\ \hspace{6.5mm}\thelinenumber}}
% \linenumbers
\pagestyle{headings}
\mainmatter

\title{Polarization Human Shape and Pose Dataset} % Replace with your title

% \titlerunning{Polarization Human Shape and Pose Dataset (PHSPD)}
% \authorrunning{Polarization Human Shape and Pose Dataset (PHSPD)}
% \author{Shihao Zou, Xinxin Zuo, Yiming Qian, Sen Wang, Minglun Gong, Li Cheng}
% \institute{University of Alberta}

% CAMERA READY SUBMISSION
\titlerunning{Polarization Human Shape and Pose Dataset}
% If the paper title is too long for the running head, you can set
% an abbreviated paper title here
%
\author{Shihao Zou\inst{1} \and
Xinxin Zuo\inst{1} \and
Yiming Qian\inst{2} \and
Sen Wang\inst{1} \and
Chuan Guo\inst{1} \and
Chi Xu\inst{3} \and
Minglun Gong\inst{4} \and
Li Cheng\inst{1}}
\authorrunning{S. Zou et al.}
% First names are abbreviated in the running head.
% If there are more than two authors, 'et al.' is used.
%
\institute{University of Alberta \and Simon Fraser University \and School of Automation, China University of Geosciences, Wuhan 430074, China \and University of Guelph\\
\email{\{szou2,xzuo,sen9,cguo2,lcheng5\}@ualberta.ca,yimingq@sfu.ca,\\xuchi@cug.edu.cn,minglun@uoguelph.ca}}
%******************

\maketitle
\section{Introduction}
Polarization images are known to be able to capture polarized reflected lights that preserve rich geometric cues of an object, which has motivated its recent applications in reconstructing detailed surface normal of the objects of interest. Meanwhile, inspired by the recent breakthroughs in human shape estimation from a single color image, we attempt to investigate the new question of whether the geometric cues from polarization camera  could be leveraged in estimating detailed human body shapes. This has led to the curation of Polarization Human Shape and Pose Dataset (PHSPD)\footnote{Our PHSPD dataset will be released soon for academic purpose only.}, our home-grown polarization image dataset of various human shapes and poses.

Our PHSPD dataset synchronizes four cameras, one polarization camera and three Kinects v2 in three different views (each  Kinect v2 has a depth and a color camera). The depth and color images from three-view Kinects v2 are used to get more accurate annotations of shape and pose in 3D space. We propose an economic yet effective approach to annotating shape and pose in 3D space. Compared with Human3.6M~\cite{human36m} that uses expensive Motion Capture system to annotate human poses, we do not require subjects to wear special tight clothes  and a lot of sensors, which makes the acquired images restrictive and impractical.

\begin{figure}[]
    \centering
    \includegraphics[width=\columnwidth]{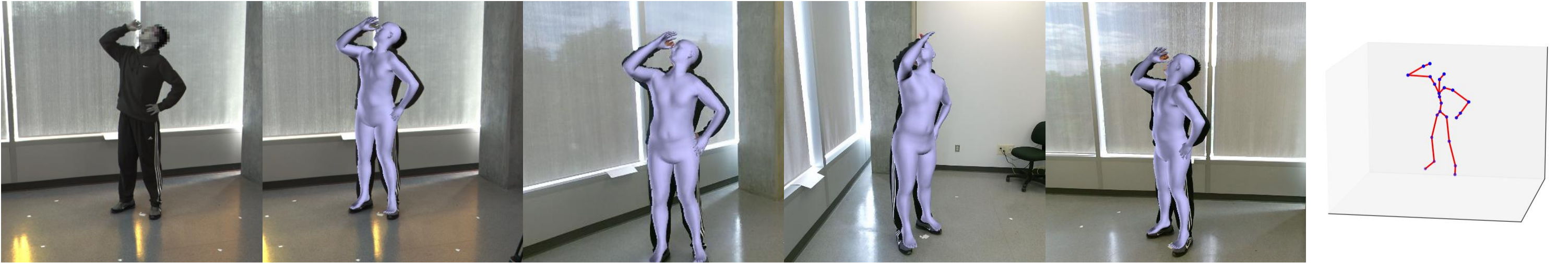}
    \includegraphics[width=\columnwidth]{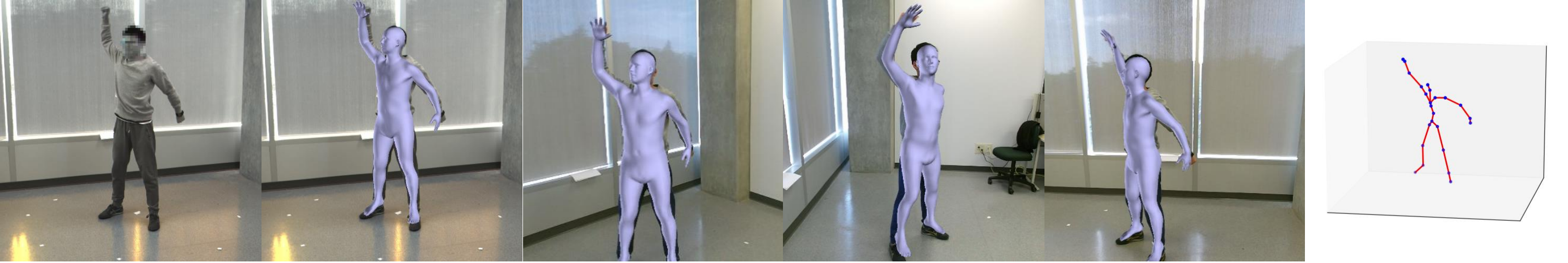}
    \includegraphics[width=\columnwidth]{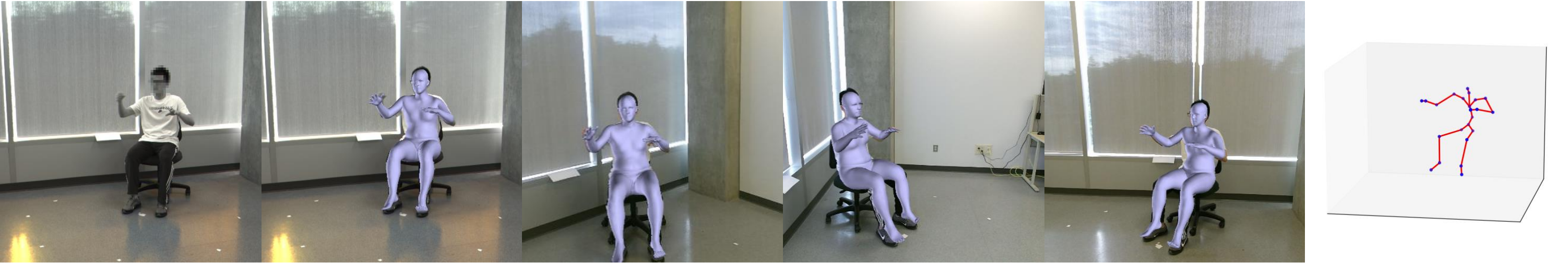}
    \includegraphics[width=\columnwidth]{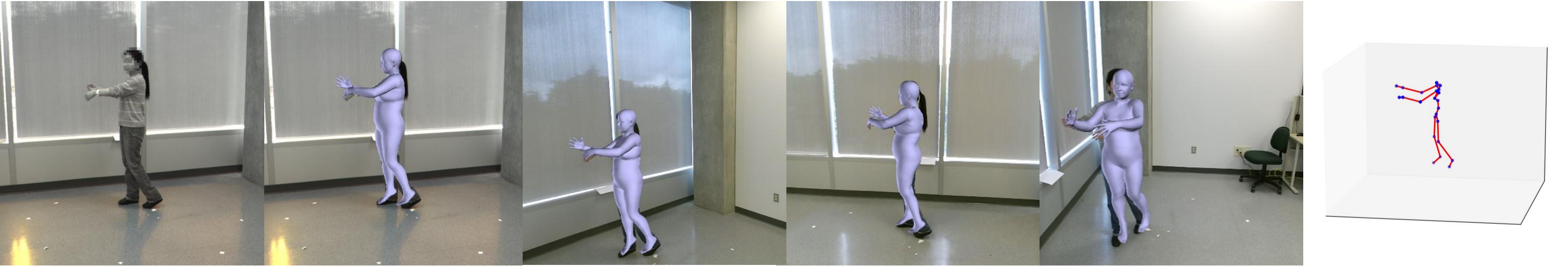}
    \includegraphics[width=\columnwidth]{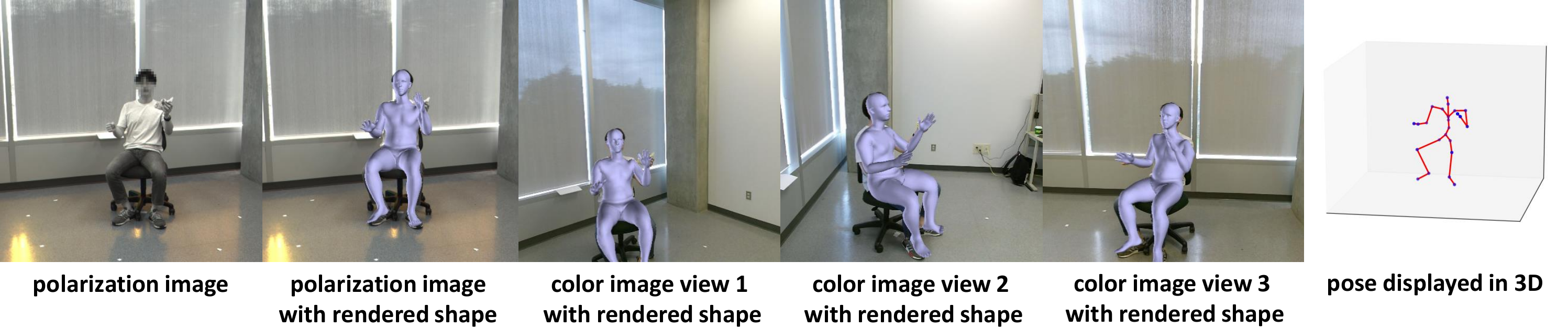}
    \caption{The figure shows our annotated shapes and poses. The first column is the polarization image for reference. The second to the fifth columns show the annotated shape rendered on the polarization image and three-view color images. The sixth column shows the annotated pose in 3D space.}
    \label{fig:multiview-demo}
\end{figure}

We show some of our annotated shapes and poses in Fig.~\ref{fig:multiview-demo}, where the shapes are rendered on the image plane and the poses are shown in 3D space. We can see that our annotated shapes and poses align well with the subjects in the image plane from four camera views.

\section{Data Acquisition} 
\begin{figure}[]
    \centering
    \includegraphics[width=0.39\columnwidth]{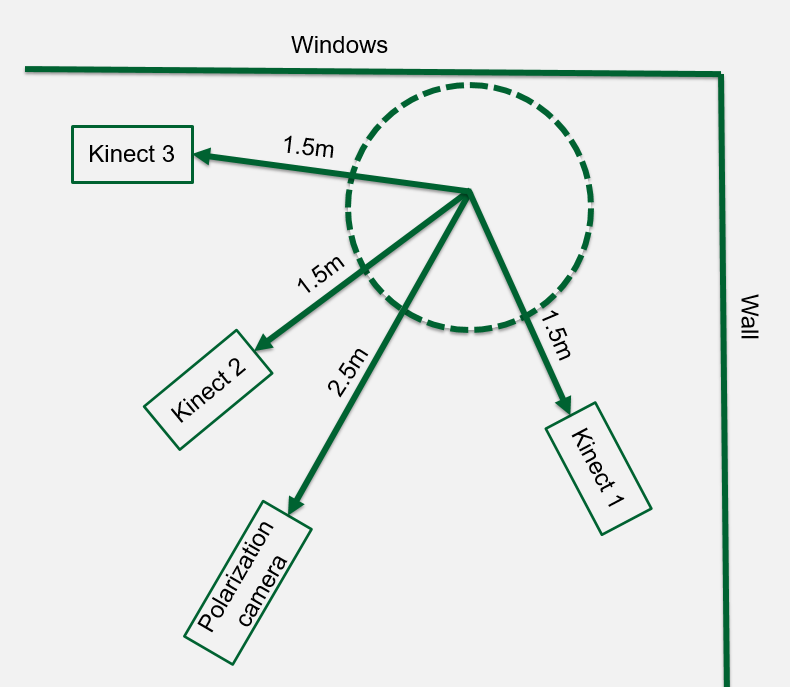}
    \includegraphics[width=0.59\columnwidth]{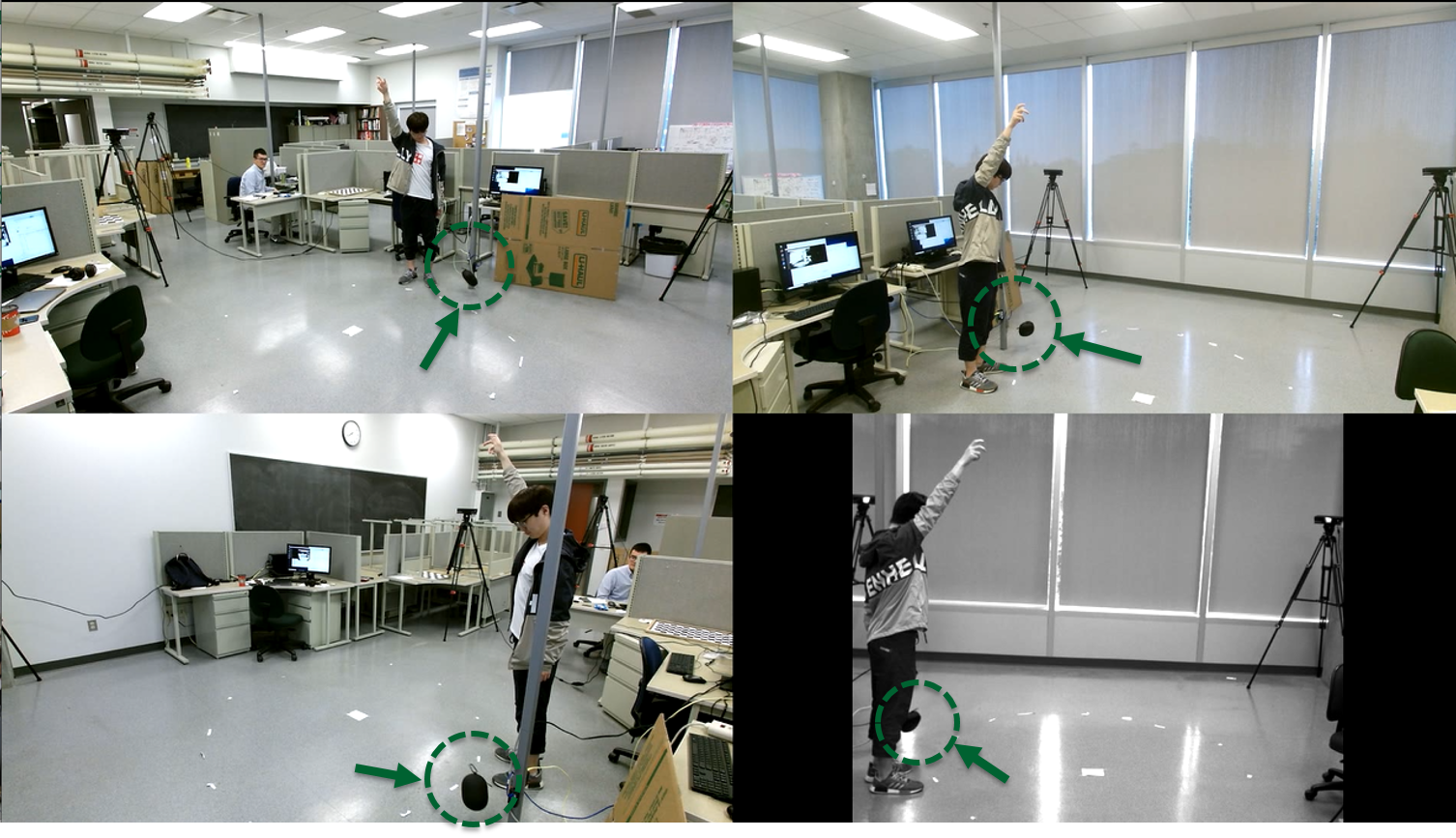}
    \caption{Left figure: the layout of our multi-camera system. Three Kinects are placed around a circle of motion area with one polarization camera. Right figure: the synchronization result of our multi-camera system. The same frame of the three-view color images and one-view polarization image are displayed. Note that the layout of our multi-camera system has been changed to the left figure, but other settings are the same.}
    \label{fig:camera-layout}
\end{figure}

Our acquisition system synchronizes four cameras, one polarization camera and three Kinects V2 in three different views (each Kinect v2 has a depth and a color camera). The layout is shown in Fig.~\ref{fig:camera-layout}. The main task in data acquisition is multi-camera synchronization. As one desktop can only control one Kinect v2, we develop a soft synchronization method. Specifically, each camera was connected with a desktop (the desktop with the polarization camera is the master and the other three ones with three Kinects are clients). We use socket to send message to each desktop. After receiving certain message, each client will capture the most recent frame from the Kinect into the desktop memory. At the same time, the master desktop sends a software trigger to the polarization camera to capture one frame into the  buffer. Practically, our synchronization system can be as fast as 15 frames per second (fps). Fig.~\ref{fig:camera-layout} shows the synchronization performance of the system that we develop. We let a bag fall down and compare the position of the bag in the same frame from four views. We can find that the positions of the bag captured by four cameras are almost the same in terms of its distance to the ground.

\begin{table}[]
    \centering
    \begin{tabular}{|c|c|}
    \toprule
        group \# & actions \\
    \midrule
        1 &  warming-up, walking, running, jumping, drinking, lifting dumbbells\\
        2 &  sitting, eating, driving, reading, phoning, waiting\\
        3 &  presenting, boxing, posing, throwing, greeting, hugging, shaking hands\\
    \bottomrule
    \end{tabular}
    \caption{The table displays the actions in each group. Subjects are required to do each group of actions for four times, but the order of the actions each time is random.}
    \label{tab:dataset-action}
\end{table}
\begin{table}[]
    \centering
    \begin{tabular}{ccccc}
    \toprule
        \makecell[c]{\ subject\ \\\ \#\ } & \ gender\  & \makecell[c]{\ \ \# of original\ \ \\\  frames\ } & \makecell[c]{\ \ \# of annotated\ \ \\\  frames\ } & \makecell[c]{\ \ \# of discarded\ \ \\\  frames\ } \\
    \midrule
        1 & female & 22561 & 22241 & 320 (1.4\%) \\
        2 & male & 24325 & 24186 & 139 (0.5\%) \\
        3 & male & 23918 & 23470 & 448 (1.8\%)\\
        4 & male & 24242 & 23906 & 336 (1.4\%)\\
        5 & male & 24823 & 23430 & 1393 (5.6\%)\\
        6 & male & 24032 & 23523 & 509 (2.1\%)\\
        7 & female & 22598 & 22362 & 236 (1.0\%)\\
        8 & male & 23965 & 23459 & 506 (2.1\%)\\
        9 & male & 24712 & 24556 & 156 (0.6\%)\\
        10 & female & 24040 & 23581 & 459 (1.9\%) \\
        11 & male & 24303 & 23795 & 508 (2.1\%)\\
        12 & male & 24355 & 23603 & 752 (3.1\%)\\
        \midrule
        total & - & 287874 & 282112 & 5762 (2.0\%)\\
    \bottomrule
    \end{tabular}
    \caption{The table shows the detail number of frames for each subject and also the number of frames that have SMPL shape and 3D joint annotations.}
    \label{tab:dataset-details}
\end{table}

Our dataset has 12 subjects, 9 males and 3 females. Each subject is required to do 3 different groups of actions (18 different actions in total) for 4 times plus one free-style group. Details are shown in Tab.~\ref{tab:dataset-action}. So each subject has 13 short videos and the total number of frames for each subject is around 22K. Overall, our dataset has 287K frames with each frame including one polarization image, three color and three depth images. Quantitative details of our dataset are shown in Tab.~\ref{tab:dataset-details}.

\section{Annotation Process}
\subsection{Shape and Pose Representation}
We represent the 3D human body shape (mesh) using SMPL model \cite{loper2015smpl}, which is a differentiable function $\mathcal{M}(\boldsymbol{\beta}, \boldsymbol{\theta})\in\mathbb{R}^{6890\times3}$ that outputs a triangular mesh with 6890 vertices given 82 parameters $[\boldsymbol{\beta}, \boldsymbol{\theta}]$. The shape parameter $\boldsymbol{\beta}\in\mathbb{R}^{10}$ is the linear coefficients of a PCA shape space that mainly determines individual body features such height, weight and body proportions. The shape space is learned from a large dataset of body scans \cite{loper2015smpl}. The pose parameter $\boldsymbol{\theta}\in\mathbb{R}^{72}$ mainly describes the articulated pose, which consists of one global rotation of the body and the relative rotations of 23 joints in axis-angle representation. The final body mesh is produced by first applying shape-dependent and pose-dependent deformations to the template body, then using forward-kinematics to articulate the body and finally deforming the surface with linear blend skinning. The 3D joint positions $\mathbf{J}\in\mathbb{R}^{24\times3}$ can be obtained by linear regression from the output mesh vertices. 

\subsection{Annotation of Shape and Pose}
The reason that we use multi-camera system to acquire image data is that multi-camera system provides much more information than a single-camera system. So the annotation of human shape and pose in 3D is more reliable.

After camera calibration and plane segmentation of human body in depth images, we have a point cloud of human surface fused from three-view depth image, and also noisy 3D pose by Kinect SDK at hand. The annotation of SMPL human shape and 3D joint position has three main steps as follows.

\subsubsection{Initial guess of 3D pose}
As the 3D pose given by Kinect SDK is noisy, we use the predicted 2D pose by OpenPose~\cite{openpose} as the criterion to decide which joint position given by Kinect SDK is correct. We select 14 aligned joints that both Openpose and Kinect have. For joint $i$ in view $j=\{1,2,3\}$, 2D joint position by OpenPose is denoted by $(v_{ij}^o, u_{ij}^o)$, and 3D joint position by Kinect by $(x_{ij}, y_{ij}, z_{ij})$ and its projected 2D joint position by $(v_{ij}^k, u_{ij}^k)$. Since we cannot figure out which joint is detected correctly by Kinect, we use the joint position by OpenPose as the criterion to decide whether this joint is correctly estimated by Kinect, that is
\begin{equation}
    w_{ij} =
    \begin{cases}
        1 & \text{if}\ \sqrt{(v_{ij}^k - v_{ij}^o)^2 + (u_{ij}^k - u_{ij}^o)^2} < 50,\\
        0 & \text{otherwise},
    \end{cases}
\end{equation}
where $50$ means the pixel distance. Then, we get the initial guess of the 3D joint position $(\hat x_i, \hat y_i, \hat z_i)$ of joint $i$ by averaging the valid positions given by three-view Kinects as
\begin{equation}
    \hat x_i = \frac{\sum_{j=1}^3 w_{ij}x_{ij}}{\sum_{j=1}^3 w_{ij}}, \hat y_i =  \frac{\sum_{j=1}^3 w_{ij}y_{ij}}{\sum_{j=1}^3 w_{ij}}, \hat z_i =  \frac{\sum_{j=1}^3 w_{ij}z_{ij}}{\sum_{j=1}^3 w_{ij}}.
\end{equation}
If none of the three-view joint positions by Kinect is correct, we consider it as a missing joint. We discard the frame with more than 2 joints missing (14 in total).

\subsubsection{Fitting shape to pose} The next step is similar to SMPLify \cite{bogo2016keep}, but instead of fitting to the 2D joints which have inherent depth ambiguity, we fit SMPL model to the initial guess of 3D pose.

\subsubsection{Fine-tuning shape to the point cloud} The final step is fine-tuning the shape to the point cloud of human surface so that the annotated SMPL shape parameters are more accurate. We iteratively optimize SMPL parameters by minimizing the distance between vertices of SMPL shape to their nearest point. Finally, we have the annotated SMPL shape parameters and 3D pose. 

Besides, we render the boundary of SMPL shape on the image to get the mask of background, and calculate the target normal using three depth images based on \cite{qi2018geonet}. Although the target normal is noisy, our experiment result shows our model can still learn to predict good and smooth normal maps.

\begin{figure}[]
    \centering
    \includegraphics[width=0.9\columnwidth]{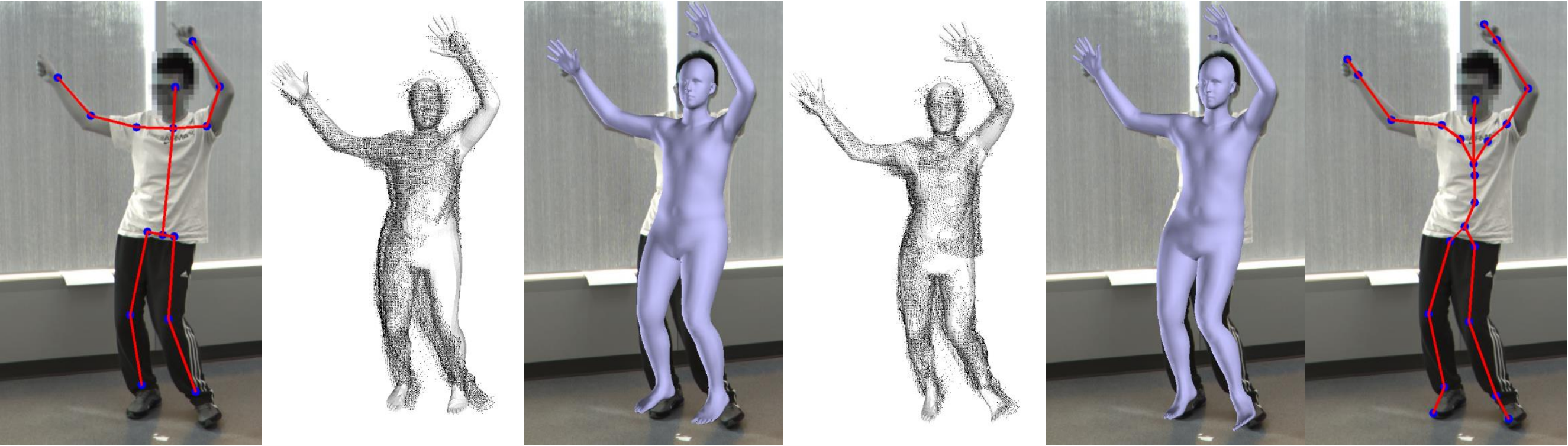}
    \includegraphics[width=0.9\columnwidth]{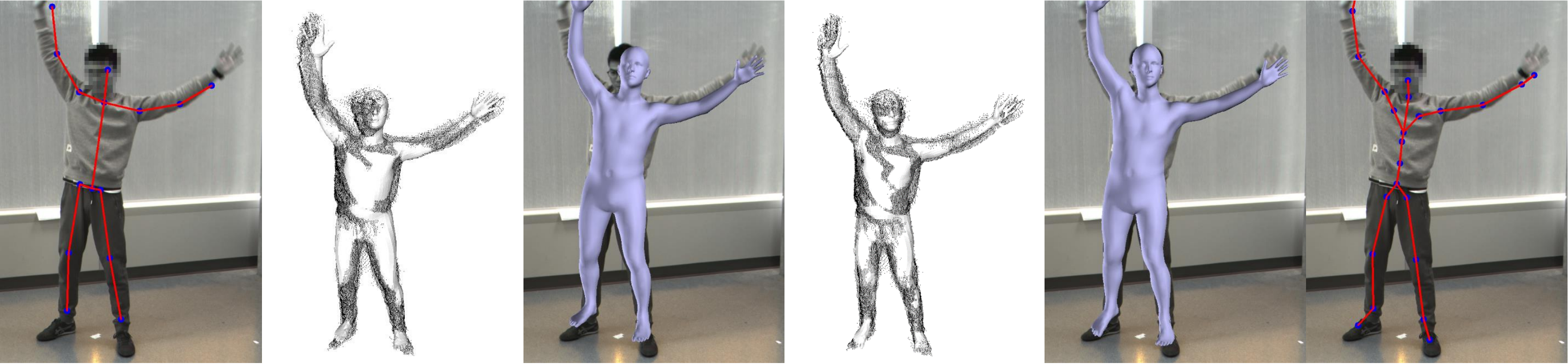}
    \includegraphics[width=0.9\columnwidth]{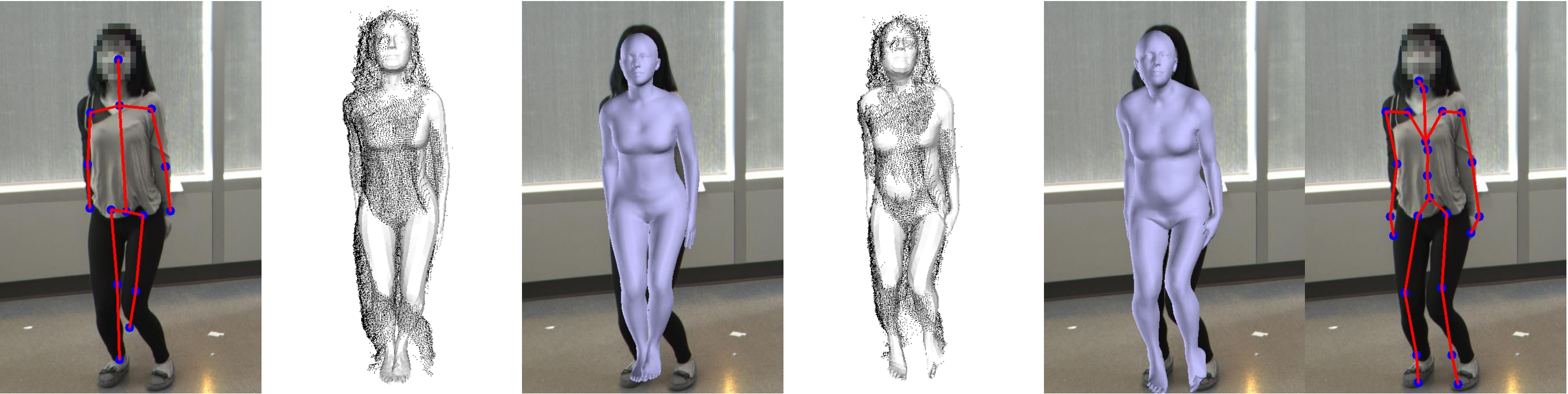}
    \includegraphics[width=0.9\columnwidth]{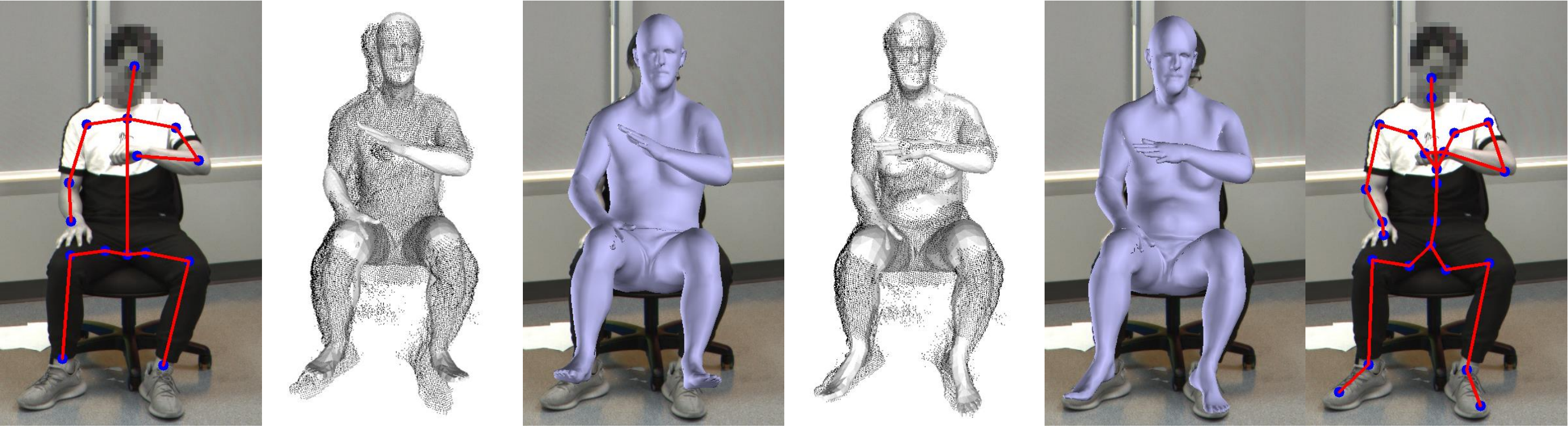}
    \includegraphics[width=0.9\columnwidth]{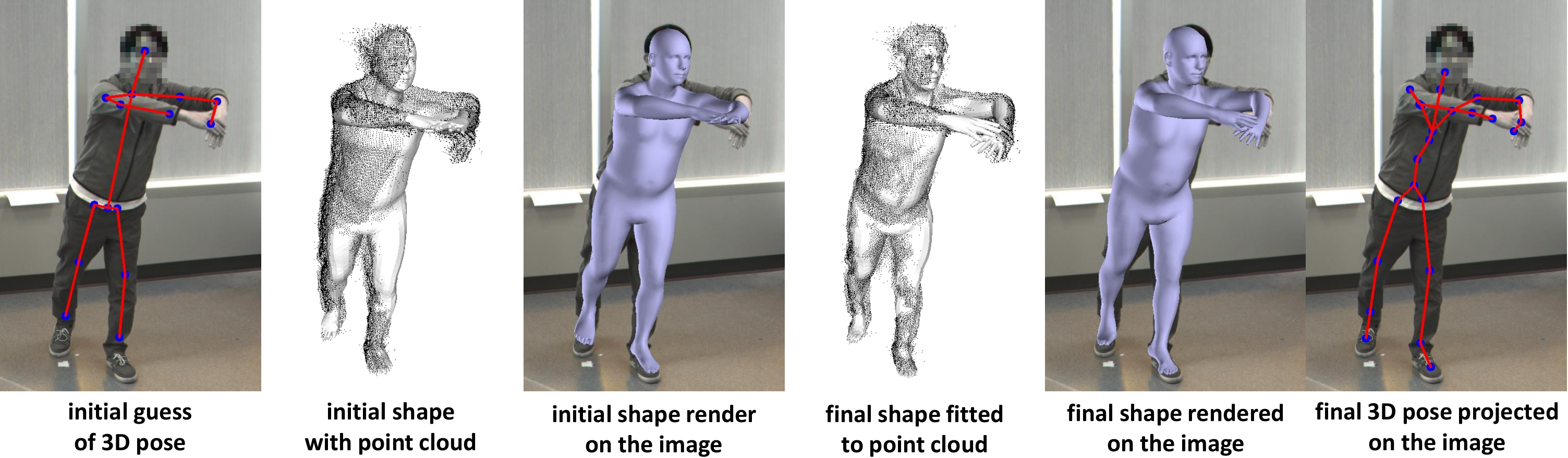}
    \caption{The figure shows our three-step annotation process. The first column shows the initial guess of 3D pose, which is projected on the polarization image. After fitting the SMPL shape to the initial pose, we show the initial fitted shape with the point cloud of human surface (black points) in the second column and the rendered shape on the image in the third column. The fourth and fifth columns show the annotated shape after fine-tuning the shape to the point cloud of human surface. The sixth column shows the corresponding annotated 3D pose projected on the polarization image.}
    \label{fig:annoataion-demo}
\end{figure}

The annotation process is shown in Fig.~\ref{fig:annoataion-demo}. Starting from the initial guess of 3D pose, we fit SMPL shape to the initial 3D pose and further fine-tune to the point cloud of human surface. Finally, we get the annotated human shape and pose for each frame. We can find from Fig.~\ref{fig:annoataion-demo} that the third step is critical to make the annotated shape align better to the subject in the image in that the pint cloud of human surface gives much more information than a skeleton-based pose. So the third step can adjust the shape to improve the alignment of body parts. Besides, we also show our annotated shape on multi-view images (one polarization image and three-view color image) and the human pose in 3D coordinate space in Fig~\ref{fig:multiview-demo}. 

\subsection{Annotation of Actions}
We select a subset of dataset PHSPD, and cut the long pose sequences into smaller pieces which aligns with prescribed action types, named as HumanAct12. The statistics of our action annotations is given in Tab.~\ref{tab:DN}, where there are 1061 motion clips which are categorized into 12 action classes and 34 sub-classes. 

\begin{table}[]
  \caption{Statistics of dataset HumanAct12.}
  \centering
  \label{tab:DN}
  \begin{tabular}{c c c c}
    \toprule
                Coarse-grained Label & Fine-grained Label & \# of Motions & \# Total \\
    \midrule
    \multirow{7}{*}{Warm up} & Warm\_up\_wristankle & 25 & \multirow{7}{*}{201} \\
                & Warm\_up\_pectoral & 45 \\
                & Warm\_up\_eblowback & 39 \\
                & Warm\_up\_bodylean\_right\_arm & 25\\
                & Warm\_up\_bodylean\_left\_arm & 24\\
                & Warm\_up\_bow\_right & 22 \\
                & Warm\_up\_bow\_left & 21 \\
    \midrule
    \multirow{1}{*}{Walk} & Walk & 43 & \multirow{1}{*}{43} \\
    \midrule
    \multirow{1}{*}{Run} & Run & 44 & \multirow{1}{*}{44} \\
    \midrule
    \multirow{2}{*}{Jump} & Jump\_handsup & 50 & \multirow{2}{*}{85} \\
                & Jump\_vertical & 35\\
    \midrule
    \multirow{5}{*}{Drink} & Drink\_bottle\_righthand & 25 & \multirow{5}{*}{81} \\
                & Drink\_bottle\_lefthand & 39 \\ 
                & Drink\_cup\_righthand & 10 \\
                & Drink\_cup\_lefthand & 3\\
                & Drink\_both\_hands & 4 \\
    \midrule
    \multirow{5}{*}{Lift\_dumbbell} & Lift\_dumbbell\_righthand & 41 & \multirow{5}{*}{198} \\
                & Lift\_dumbbell\_lefthand & 41 \\ 
                & Lift\_dumbbell\_bothhands & 43 \\
                & Lift\_dumbbell\_overhead & 39\\
                & Lift\_dumbbell\_bothhands\_bend\_legs & 34 \\
    \midrule
    \multirow{1}{*}{Sit} & Sit & 47 & \multirow{1}{*}{47} \\
    \midrule
    \multirow{3}{*}{Eat} & Eat\_righthand & 27 & \multirow{3}{*}{68} \\
                & Eat\_lefthand & 18 \\ 
                & Eat\_pie/burger & 23 \\
    \midrule
    \multirow{1}{*}{Turn\_steering\_wheel} & Turn\_steering\_wheel & 46 & \multirow{1}{*}{46} \\
    \midrule
    \multirow{2}{*}{Phone} & Take out phone, call and put back & 19 & \multirow{2}{*}{52} \\
                & Call with left hand & 33 \\ 
    \midrule
    \multirow{4}{*}{Boxing} & Boxing\_left\_right & 21 & \multirow{4}{*}{116} \\
                & Boxing\_left\_upwards & 34 \\
                & Boxing\_right\_upwards & 37 \\
                & Boxing\_right\_left & 24 \\
    \midrule
    \multirow{2}{*}{Throw} & Throw\_right\_hand & 47 & \multirow{2}{*}{80} \\
                & Throw\_both\_hand & 33 \\
    \midrule
    \multirow{1}{*}{\textbf{Entire Dataset}} & - & - & \multirow{1}{*}{\textbf{1061}} \\
    \bottomrule
  \end{tabular}
\end{table}

\newpage
\bibliographystyle{splncs}
\bibliography{egbib}
\end{document}